# A DeepSeek-Powered AI System for Automated Chest Radiograph Interpretation in Clinical Practice

Yaowei Bai[1,2,3]*, Ruiheng Zhang[1,4]*, Yu Lei[2,3]*, Xuhua Duan[5]*, Jingfeng Yao[6], Shuguang Ju[5], Chaoyang Wang[7], Wei Yao[2,3], Yiwan Guo[2,3], Guilin Zhang[2,3], Chao Wan[8], Qian Yuan[9], Lei Chen[2,3], Wenjuan Tang[2,3], Biqiang Zhu[10], Xinggang Wang[6], Tao Sun[11], Wei Zhou[12], Dacheng Tao[13#], Yongchao Xu[1,4#], Chuansheng Zheng[2,3#], Huangxuan Zhao[1,2,4#], Bo Du[1,4#]

Affiliations:

[1] School of Computer Science, Wuhan University, Wuhan, 430072, China

[2] Department of Radiology, Union Hospital, Tongji Medical College, Huazhong University of Science and Technology, Wuhan, 430022, China

[3] Hubei Provincial Clinical Research Center for Precision Radiology & Interventional Medicine, Wuhan 430022, China

[4] National Engineering Research Center for Multimedia Software and Hubei Key Laboratory of Multimedia and Network Communication Engineering, Wuhan University, Wuhan 430072, China

[5] Department of Interventional Radiology, The First Affiliated Hospital of Zhengzhou University, Zhengzhou, 450000, China

[6] School of Electronic Information and Communications, Huazhong University of Science and Technology, Wuhan, 430074, China

[7] Department of Interventional Radiology, The First Affiliated Hospital of Henan University of Science and Technology, Luoyang, 471003, China

[8] Cancer Center, Union Hospital, Tongji Medical College, Huazhong University of Science and Technology, Wuhan, 430022, China

[9] Department of Nephrology, Union Hospital, Tongji Medical College, Huazhong University of Science and Technology, Wuhan, 430022, China

[10] Department of Urology, The Central Hospital of Wuhan, Tongji Medical College, Huazhong University of Science and Technology, Wuhan, 430014, China

[11] Department of Interventional Radiology, The First Affiliated Hospital of University of Science and Technology of China, Hefei, 230001, China

[12] Wuhan Artificial Intelligence Computing Center, Wuhan Supercomputing Center, Wuhan, 430072, China

[13] School of Computer Science and Engineering, Nanyang Technological University, Singapore, 639798, Singapore

*Contributed equally to the study;

#Corresponding authors. Email: dacheng.tao@ntu.edu.sg; yongchao.xu@whu.edu.cn; hqzcsxh@sina.com; zhao_huangxuan@sina.com; dubo@whu.edu.cn

**Abstract**

A global shortage of radiologists has been exacerbated by the significant volume of chest X-ray workloads, particularly in primary care. Although multimodal large language models show promise, existing evaluations predominantly rely on automated metrics or retrospective analyses, lacking rigorous prospective clinical validation. Janus-Pro-CXR (1B), a chest X-ray interpretation system based on DeepSeek's Janus-Pro model, was developed and rigorously validated through a multicenter prospective trial (NCT07117266). Our system outperforms state-of-the-art X-ray report generation models in automated report generation, surpassing even larger-scale models including ChatGPT 4o (200B parameters), while demonstrating reliable detection of six clinically critical radiographic findings (area under the curve, AUC > 0.8). Retrospective evaluation confirms significantly higher report accuracy than Janus-Pro and ChatGPT 4o. In prospective clinical deployment, AI assistance significantly improved report quality scores (4.36±0.50 vs. 4.12±0.80, P < 0.001), reduced interpretation time by 18.3% (P < 0.001), and was preferred by a majority of experts (≥3 out of 5) in 54.3% of cases. Through lightweight architecture and domain-specific optimization, Janus-Pro-CXR improves diagnostic reliability and workflow efficiency, particularly in resource-constrained settings. The model architecture and implementation framework will be open-sourced to facilitate the clinical translation of AI-assisted radiology solutions.

**Introduction**

The global shortage of radiologists presents a critical challenge, as in most regions, the expansion of imaging applications in patient care has outpaced the capacity of radiologists to meet the demand[1,2]. This issue is particularly pronounced in low-income regions, where countries report just 1.9 radiologists per million residents, compared to 97.9 per million in high-income countries[3,4]. Chest X-ray (CXR), the most fundamental and widely utilized imaging modality, remains indispensable in clinical practices such as detecting pulmonary infections and screening for tumors. X-ray diagnostics contribute significantly to the workload of radiologists, especially in primary healthcare settings[5].

Recent advances in artificial intelligence (AI) demonstrate substantial potential to enhance diagnostic efficiency, optimize medical resource allocation, and maintain the quality of healthcare[6]. Currently, most AI applications focus on classifying and quantitatively analyzing imaging features for specific diseases, but clinical imaging diagnoses are far more complex than simple classification tasks[7]. Developing an intelligent imaging system with automated report generation at its core could be crucial in improving diagnostic efficiency and alleviating the strain on radiologists[8].

Despite the growing body of research on CXR report generation, most existing models are built from scratch[9-11], presenting inherent limitations such as low data efficiency, modal fragmentation, knowledge transfer challenges, and an exacerbation of the long-tail problem[12]. Transfer learning, leveraging pre-trained knowledge and cross-modal alignment, could significantly improve the accuracy and clinical relevance of report-generation systems[8,13]. However, many current models are either prohibitively large, underperforming, or non-open-source, limiting clinical applicability. Additionally, most prior studies have evaluated generated reports using natural language generation metrics, without assessing the actual clinical impact[14-17]. Though some studies have provided comprehensive evaluations[12,18,19], they have largely relied on retrospective data in collaboration scenarios between clinicians and AI, without prospective validation in real clinical settings. Consequently, the clinical value of multimodal large models in chest radiograph interpretation remains uncertain. The recently released open-source multimodal large language model Janus-Pro by DeepSeek[20,21], with its combination of high performance and low cost, offers a new pathway for developing medical-specific report-generation systems. However, the application of this model in medical imaging has not been systematically tested, and existing general multimodal models lack task-specific optimization, necessitating further fine-tuning.

To address these gaps, this study introduces Janus-Pro-CXR, a lightweight CXR-specific model (Figure 1) developed from the unified Janus-Pro model and the public MIMIC-CXR[22] and CheXpert Plus[23] medical imaging datasets through supervised fine-tuning. With 1 billion parameters, the model achieves rapid imaging analysis with a latency of 1-2 seconds on a laptop equipped with a GeForce RTX 4060 (8GB). Its low fine-tuning cost further supports deployment in regions with limited medical resources. The model's performance in core tasks, including disease diagnosis and report generation, was rigorously evaluated using multicenter retrospective data from 27 hospitals. A multicenter prospective verification scheme was also implemented for clinical collaboration scenarios. The lightweight architecture and domain-specific optimization of Janus-Pro-CXR enhanced diagnostic accuracy and workflow efficiency, offering particular benefits for radiologists and settings with limited resources. The model architecture will be open-sourced to facilitate the clinical translation of AI-assisted radiology solutions.

**Results**

**Patients**

The general large language model, Janus-Pro, underwent supervised fine-tuning using the MIMIC-CXR, CheXpert Plus and CXR-27 datasets (multicenter retrospective data) (Figure 2). In the retrospective study, 384,208 images were allocated for the first two stages of model fine-tuning from the MIMIC-CXR and CheXpert Plus datasets and 11,156 images from 27 hospitals in China (CXR-27 dataset) were used for supervised fine-tuning. The remaining data from the CXR-27 dataset (n = 1,240) and a portion of the MIMIC-CXR dataset (n = 2,365) were used to assess model performance. For the prospective study, data were sourced from three hospitals in China, with a total of 296 patients enrolled in the AI-radiologist collaboration. Baseline patient data are provided in Supplementary Tables 1 and 2. The core conclusions of this study are grounded in the findings of the prospective study, while the retrospective study merely provides methodological scaffolding and supporting findings.

**Primary outcomes**

The primary outcomes of this study included report quality scores, pairwise preference tests, agreement evaluations, and reading time for the radiology reports produced by the two groups of junior radiologists in the prospective study. In the prospective study, junior radiologists in the AI-assisted group used reports generated by Janus-Pro-CXR as references and modified the content as needed, while radiologists in the standard care group independently drafted their reports. For each patient's imaging data, two junior radiologists from different groups independently reviewed the scans and generated separate reports. These reports were then reviewed, revised, and finalized by a senior radiologist before being issued for clinical use.

This prospective study demonstrated that AI significantly enhanced the quality of radiology reports created by junior radiologists (Table 1). Inter-rater reliability verification in the prospective study is presented in Supplementary Table 3. In terms of report quality, the AI-assisted group achieved a mean score of $4.36\pm0.50$, significantly higher than the standard care group's score of $4.12\pm0.80$ (Mean of differences = 0.25, $P < 0.001$, 95% CI = 0.216-0.283) (Figure 3A). The agreement score of the AI-assisted group's reports, assessed using the RADPEER scoring system, was $4.30\pm0.57$, notably higher than the standard care group's score of $4.14\pm0.84$ (Mean of differences = 0.16, $P < 0.001$, 95% CI = 0.119-0.200) (Figure 3B). In the pairwise preference test between radiologists and AI, 30.4% of the reports generated by AI were preferred by three or more evaluation experts. When comparing radiologists with and without AI assistance, 54.3% of AI-assisted reports were favored by three or more experts (Figure 3C, 3D, and 3E). An example illustrating the improvement in report quality through AI assistance is presented in Supplementary Table 4. The analysis of work efficiency showed that the AI-assisted group wrote reports in an average of 120.6 ± 45.6 seconds, significantly faster than the standard care group, which took 147.6 ± 51.1 seconds (Mean of differences = 27.0, 18.3% reduction, P < 0.001, 95% CI = 19.2-34.8) (Figure 3F). Subgroup analysis revealed that for complex cases (those with ≥ 3 imaging findings), the time advantage for the AI-assisted group remained substantial (165.1 ± 29.4 vs. 197.6 ± 26.9 seconds, Mean of differences = 32.5s, 16.4% reduction, P < 0.001, 95% CI = 31.6-33.4).

**Second outcomes**

In the retrospective study, comparative analyses between AI-generated reports and original reports were performed, report quality scores, pairwise preference tests, and agreement evaluations as secondary outcomes.

Among the 1,240 cases in the CXR-27 test set, 300 images were randomly selected for subjective evaluation. Inter-rater reliability verification in the retrospective study is presented in Supplementary Table 5. First, a confusion matrix was used to assess the language professionalism and standardization of the reports generated by the large language models, without considering content correctness. The confusion matrix results revealed that the reports generated by Janus-Pro-CXR closely resembled the style of published reports, making it difficult for evaluators to distinguish them from the original reports. In contrast, reports generated by Janus-Pro and ChatGPT 4o were easily identifiable (Figure 3G and Supplementary Figure 1). When comparing the reports generated by Janus-Pro-CXR, Janus-Pro, and ChatGPT 4o to the published reports, the evaluation team assigned a quality score of 3.22±1.14 to the Janus-Pro-CXR-generated reports. This was significantly higher than the scores for reports generated by Janus-Pro (1.57±0.63) and ChatGPT 4o (1.70±0.76) (Figure 3H). The agreement scores between the Janus-Pro-CXR-generated reports and the published reports reached 3.10±1.05, higher than the agreement scores for Janus-Pro (1.66±0.60) and ChatGPT 4o (1.74±0.75) (Figure 3I). In the pairwise preference test, 15.3% of the reports generated by Janus-Pro-CXR were favored by three or more evaluation experts, significantly surpassing the proportions for Janus-Pro (2.8%) and ChatGPT 4o (5.2%) (Figure 3J).

**Performance of Janus-Pro-CXR in the retrospective study**

We propose the Janus-Pro-CXR system, which employs a large-small model collaborative framework: it injects diagnostic results via an expert model, integrates multimodal clinical data through a unified model, and explicitly models clinical image reading logic (Figure 4A). In the assessment of automated report generation metrics, Janus-Pro-CXR demonstrated superior performance across multiple dimensions (Figure 4B and 4C, Supplementary Table 6). Evaluation on the MIMIC-CXR test set highlighted the model's ability to achieve balanced performance in recognizing both common and rare diseases. Our model ranks first among all models in terms of Micro-avg F1-5 (63.4) and Macro-avg F1-5 (55.1), and second in terms of Micro-avg F1-14 (59.9) and Macro-avg F1-14 (42.3) in the MIMIC-CXR test set. Additionally, the model's Radiology Graph (RadGraph) F1 score for the medical entity relationship extraction task reached 25.8, surpassing that of other models for similar tasks. This advantage was further evident in the CXR-27 test set, where Janus-Pro-CXR excelled in natural language generation metrics, ranking first in all indicators, with its RadGraph F1 score reaching as high as 58.6. Additionally, Janus-Pro-CXR-Zero also performed well in the CXR-27 test set, ranking second in most indicators.

To assess consistency with the reference standard, Cohen's kappa coefficient was employed. The model showed high consistency with the reference standard in classifying findings such as support devices and pleural effusion in the CXR-27 test set (Figure 4D). Diagnostic performance was also strong (Figure 4E, Supplementary Figure 2, and Supplementary Table 7), with AUC values > 0.8 for six key findings, including support devices (AUC = 0.931), pleural effusion (AUC = 0.931), and pneumothorax (AUC = 0.921) in the CXR-27 test set. However, the model's ability to recognize the complex or subtle findings, such as fractures, requires further improvement.

**Validation of Multi-Image Input Capability of the Model**

In order to verify the model's multi-image input capability, some cases (n=50) in the test set containing historical chest X-rays and cases (n=50) containing posteroanterior (PA) and lateral chest

X-rays were selected. The upgraded model was used for image processing and report generation, and five evaluation experts conducted subjective scoring on the generated reports. Results demonstrated that the model could stably receive and integrate the aforementioned multi-image input information, successfully completing the automatic generation of corresponding clinical reports. Concurrently, subjective evaluations of the reports generated from these multi-image inputs were conducted in accordance with the report quality scoring criteria. The results showed that the report quality score for cases involving historical chest radiographs was 3.19±1.22, while that for cases with both PA and lateral chest radiographs was 3.42±1.02. Notably, a supplementary analysis of the 50 PA-lateral cases—where reports were generated using only PA images—yielded a lower score of 3.23±1.06, indicating that simultaneous input of PA and lateral views contributes to improved report quality (P<0.001). These findings further validate the practical applicability of this function in clinical settings. One representative case with historical chest radiographs (Supplementary Table 8) and one with PA and lateral chest radiographs (Supplementary Table 9) are provided respectively for illustration. In addition, we have conducted prospective validation involving 50 patients who had posteroanterior and lateral chest X-rays (Supplementary Figure 3). Specific results demonstrated that the AI-assisted group achieved a higher report quality score than the standard care group (4.37±0.50 vs. 4.16±0.76, P<0.001); the AI-assisted group also reached a higher agreement score than the standard care group (4.38±0.58 vs. 4.22±0.90, P<0.001); and in terms of work efficiency, the average report reading time for the AI-assisted group was shorter than the standard care group (120.5±46.9 seconds vs. 153.6±57.0 seconds, P<0.001).

**Discussion**

In this study, Janus-Pro-CXR, an intelligent report generation system tailored for radiology clinical practice, was developed based on the DeepSeek multimodal large language model (Janus-Pro). The core conclusions of this study are grounded in the findings of the prospective study, while the retrospective study merely provides methodological scaffolding and supporting findings. The domain-specific optimization of Janus-Pro-CXR has yielded substantial advantages in natural language generation metrics[12,24-27] and demonstrated good performance in diagnostic accuracy and report generation quality. Prospective clinical validation confirmed that Janus-Pro-CXR not only significantly enhances the diagnostic accuracy of radiologists but also effectively improves their work efficiency. Furthermore, with just 1 billion parameters, the model achieves state-of-the-art performance in automated report generation metrics, outperforming both specialized large models in its class and higher-parameter models including ChatGPT 4o (200B)[28]. This parameter efficiency gives it a distinct advantage for deployment in resource-constrained settings.

The lightweight architecture of the model plays a critical role in facilitating the clinical application of AI-assisted diagnostic technologies[29]. In the present study, Janus-Pro-CXR, a lightweight CXR-specific model derived from the unified Janus-Pro, was introduced through supervised fine-tuning. With a compact 1 billion-parameter design, the model achieves rapid imaging analysis with a latency of 1-2 seconds on a laptop equipped with a GeForce RTX 4060 (8GB). This characteristic is crucial for practical deployment in primary healthcare settings. Unlike traditional AI models that require high-performance computing hardware, Janus-Pro-CXR maintains excellent diagnostic performance while significantly lowering the hardware requirements, making AI-assisted diagnostic services accessible in resource-constrained settings[30]. Our decision to pursue a lightweight 1-billion-parameter architecture, and employed a collaborative framework that combines a unified model with an expert classification model, yielded superior performance compared to previous models with larger parameter sizes (e.g., 3B, 7B). This outcome aligns with scaling laws[31], which posit a power-law relationship between performance gains and increases in parameters, data scale, and compute, with improvements diminishing as size grows. Recent research corroborates this, demonstrating minimal performance difference between 3B and much larger (e.g., 40B) models for certain tasks[32], suggesting 3B parameters may already be sufficient, and further scaling risks resource inefficiency. Consequently, beyond a certain threshold, performance enhancements rely less on sheer parameter count and more on superior data quality, refined architectures, and optimized training strategies. Moreover, the choice of a lightweight 1B model was driven by deployment practicality. Its significantly lower computational demands for fine-tuning and inference compared to larger models, coupled with its open-source nature, greatly facilitates dissemination to underdeveloped regions and primary care facilities[29]. Notably, models often require further fine-tuning when applied to different populations. The model requires only 10,000 images for domain adaptation fine-tuning to reach professional-level diagnostic accuracy, and its low fine-tuning cost further enhances its applicability. The lightweight design does not compromise the model's performance; rather, through domain-specific optimization, it surpasses both general and specialized large models in CXR diagnosis tasks, achieving exceptional results in key metrics, such as RadGraph score and F1 values for disease recognition.

Despite the increasing number of studies on AI-based report generation, which holds great potential for optimizing radiology workflows, there remains a lack of prospective validation of automatic report generation systems in real-world clinical scenarios. This gap may be attributed to the limited accuracy of reports generated by current systems and the absence of a widely accepted framework for evaluating clinical utility. Building upon the retrospective study that assessed the accuracy of Janus-Pro-CXR report generation, this study uniquely conducted a multicenter prospective clinical validation. The results revealed that Janus-Pro-CXR significantly enhanced the diagnostic quality of junior radiologists, improving report quality scores by 0.25 points, while also reducing the average report time by 27.0 seconds (18.3% reduction). These improvements are highly aligned with core clinical needs. Specifically, enhanced report quality can directly reduce the incidence of missed diagnoses and misdiagnoses, thereby lowering risks in clinical decision-making. After all, in medical practice, even minor biases in reports may impose a heavy medical burden on patients or even endanger their lives. Meanwhile, shorter image interpretation time can optimize radiology workflows. Considering the characteristics of clinical practice in large-scale medical centers in China, the radiologists read approximately 200 chest radiographs per person in a day if they devote the entire day solely to this task. Based on the 27 seconds saved per report in the AI-assisted group in this study, radiologists can accumulate a total of 90 minutes of savings per day. This saved time can be used not only for in-depth analysis of complex cases but also for alleviating visual fatigue caused by prolonged and high-intensity image reading, thereby reducing the risk of missed diagnoses due to fatigue and indirectly improving the safety of diagnosis and treatment.

Furthermore, the positive rate of pneumonia diagnosis in the AI-assisted group was higher than in the standard care group (52.4% vs. 36.1%, $P < 0.001$). The AI system likely enhanced the decision-making confidence of junior radiologists by offering structured descriptions of abnormal signs and diagnostic prompts. In cases of uncertainty, radiologists in the AI-assisted group were able to convert ambiguous observations (e.g., "possible bilateral bronchitis") into definitive diagnoses (e.g., "bilateral bronchitis"). This confidence boost is closely linked to the diagnostic capabilities and clinical experience of junior radiologists[33].

Generalist models like ChatGPT 4o excel in cross-domain versatility and strong natural language comprehension, with preliminary potential in medical text processing support[3]. As a chest X-ray-specific model, Janus-Pro-CXR's core advantages over generalist models such as ChatGPT 4o lie primarily in two aspects: clinical adaptability and practical deployment value. First, it demonstrates better detection accuracy for clinical signs of chest diseases. Second, in terms of terminology use, Janus-Pro-CXR strictly adheres to radiological diagnostic standards, effectively avoiding issues common in generalist models such as colloquial terminology and disorganized content structure. Third, Janus-Pro-CXR has a significantly lower deployment barrier: its lightweight architectural design eliminates the need for high-end hardware, enabling fast inference on standard workstations. This not only reduces reliance on cloud-based inference but also mitigates the risk of data privacy breaches. On the other hand, existing chest X-ray-specific models (e.g., MAIRA-2[34], CheXagent[35]) have shown great field value, with progress in key tasks and valuable experience for chest imaging AI's clinical translation. When compared with other chest X-ray-specific models, Janus-Pro-CXR achieves better performance in objective metrics related to report generation and key lesion detection. Furthermore, other chest X-ray-specific models generally have significant accessibility limitations[12,36]. For instance, some do not open-source their model weights or training codes, which makes independent deployment difficult for clinical institutions. Others require high GPU memory that standard workstation configurations cannot support. This is precisely why we did not conduct comparisons with these specialist models in our prospective study, even though they may also exhibit equally good performance in certain tasks.

AI assistance may also enhance the work efficiency of senior physicians. The generation and release of radiology diagnostic reports adhere to a stringent standardized workflow. Initially, junior radiologists perform comprehensive imaging analysis to generate preliminary reports, which subsequently undergo dual validation by senior radiologists to ensure diagnostic accuracy and protocol compliance. Upon verification, the reports are electronically signed by senior radiologists for official release, with real-time synchronization to the hospital information system for clinical reference. Notably, the AI-assisted workflow not only enhances the efficiency of junior radiologists but also potentially supports senior radiologists' review process by providing clearer diagnostic suggestions. This may reduce the need for report modifications by senior physicians and accelerate the review speed, thereby improving overall workflow efficiency.

This study has several limitations. Firstly, the model's recognition performance for complex or subtle findings, such as edema and pleural thickening, is relatively constrained, a common challenge with current multimodal large language models[37]. An example of the model's suboptimal performance is provided in Supplementary Table 10. This limitation may be due to the mild manifestations and variable morphologies of these findings, as well as insufficient sample representation of some conditions in the training and fine-tuning datasets. However, AI collaboration has still contributed to improvements in report quality and work efficiency. Future research should focus on optimizing the model's ability to recognize subtle findings and ensuring that training and fine-tuning datasets are more balanced and clinically representative.

Second, the real-world applicability of multi-image integration capability is one of the core prerequisites for the successful clinical translation of multimodal medical imaging AI[38]. Currently, Janus-Pro-CXR has completed validation for multi-image input of posteroanterior and lateral chest radiographs, the most common scenario in clinical practice. Its performance advantages have been confirmed through prospective studies. However, complete prospective validation is still lacking for the longitudinal integration capability of historical chest radiographs. This capability refers to the analysis of dynamic changes in serial follow-up images. This is mainly because cases with complete sequences of historical chest radiographs can only be obtained through long-term clinical follow-up, resulting in a relatively long accumulation cycle. It is difficult to establish a complete cohort in the short term. Subsequent research will focus on advancing the prospective validation of this function to enhance the model's clinical applicability.

Furthermore, this study still has certain limitations in aspects such as prospective clinical validation. Compared with other chest X-ray-specific models such as MAIRA-2[34] and CheXagent[35], Janus-Pro-CXR performs better in objective metrics related to report generation and key lesion detection. However, other specialized models generally have significant accessibility limitations. They may not be open-source or may require high GPU memory that standard configurations cannot support[12,36]. For this reason, we did not compare Janus-Pro-CXR with these models in the prospective study. Some of these models may still perform equally well in specific tasks. Meanwhile, the reference standard adopted in this study consists of the published radiological reports, which are not absolutely perfect "gold standards" and their accuracy is susceptible to the complexity of clinical scenarios and differences in individual radiologists' judgments. On the other hand, although Janus-Pro-CXR outperforms Janus-Pro and GPT-4o in independent report generation in retrospective analyses, it still has significant limitations. This is further confirmed by its performance in independent report generation in the prospective study. It indicates that Janus-Pro-CXR can only serve as an auxiliary tool for radiologists at the current stage and cannot replace them. In future research, we will further optimize Janus-Pro-CXR's independent report generation capability. We will also explore the workflow model where AI-generated reports are directly submitted to senior physicians for review. This aims to provide more reliable evidence-based support for improving the efficiency of clinical translation. Finally, the collaboration mode between clinicians and AI can be more intricate than what was explored in this study. The ideal form of interaction should involve two-way real-time communication, similar to that with a senior consultation expert who not only answers radiologists' questions promptly but also offers proactive suggestions for diagnostic reports, such as identifying potential misdiagnoses or missed lesions. Although recent research on multimodal interactive medical AI has laid the foundation for such collaborative diagnostic systems[39,40], much work remains to be done to develop an intelligent report assistance system that fully meets clinical needs.

This study represents the first prospective validation of Janus-Pro-CXR, highlighting the clinical value of AI-radiologist collaboration in real-world diagnostic practice. Through systematic supervised fine-tuning and comprehensive clinical validation, Janus-Pro-CXR not only improved diagnostic accuracy and workflow efficiency but also successfully transitioned from an auxiliary tool to a reliable "digital colleague." This advancement has proven particularly beneficial for radiologists and healthcare providers, especially in resource-limited settings. Although the current validation focused on CXR interpretation, the Janus-Pro-CXR framework offers a scalable paradigm with potential applications across various imaging modalities, including CT, MRI, and ultrasound. The core technological advantages of this system—its lightweight architecture, domain-specific optimization, and seamless human-AI collaborative workflow—demonstrate broad applicability across medical imaging domains.

**Methods**

**Ethics Approval and Consent**

Approval for this study was granted by the Institutional Review Boards of all participating centers, and it was duly registered with ClinicalTrials.gov (Identifier: NCT07117266). Informed consent was waived for the retrospective studies. For the prospective trial, written informed consent was obtained from all participating patients. All trials adhered to the CONSORT-AI, SPIRIT-AI, and SAGER reporting guidelines[41-44].

**Model Establishment**

The study utilized Janus-Pro (1B) as the base model, which underwent full fine-tuning. Fine-tuning was performed in three stages: the first two stages used the MIMIC-CXR and CheXpert Plus datasets to acquire proficiency in basic diagnosis and report generation, and subsequently the CXR-27 dataset was used to address data variability and adapt the writing style. The detailed process for model development is outlined in Supplementary Appendix 1. Regarding the interface of the model, relevant information can be obtained from Supplementary Figure 4.

**Trial design**

This research encompassed both retrospective and prospective components. The retrospective study focused on fine-tuning and evaluating the multimodal large language model, while the prospective study aimed to explore AI collaboration with radiologists. In the retrospective study, 384,208 images were allocated to the first-stage model fine-tuning from the MIMIC-CXR and CheXpert Plus datasets (Supplementary Appendix 2) and 11,156 images from 27 hospitals in China (CXR-27 dataset) were used for supervised fine-tuning. The remaining data from the CXR-27 dataset (n =

1,240) and a portion of the MIMIC-CXR dataset (n = 2,365) were used to assess model performance. The prospective study was conducted across three medical institutions in China (Supplementary Appendix 3). Based on the previous studies, the prospective research adopts a similar reader study design[12,35].

**Participants**

After completing data cleaning and quality screening of the MIMIC-CXR and CheXpert Plus datasets, a total of 167,496 radiographic images from the MIMIC-CXR dataset and 222,103 radiographic images from the CheXpert Plus dataset met the inclusion criteria (Supplementary Appendix 4). Among these, 384,208 images were allocated to the first-stage model fine-tuning, and 2,365 images were designated as the test set. For MIMIC-CXR, testing was strictly conducted in accordance with its official train-test split standards[45]. Specifically, the 2,365 test images in this study were directly selected from the official test set defined in the dataset's core documentation. For the retrospective study component, a total of 12,721 patients' chest radiographs were collected. After subsequent screening, 12,396 qualified images were obtained, of which 11,156 were used for the second-stage model fine-tuning, and 1,240 were retained as the final test set. The test set division for the CXR-27 dataset was performed using a random 9:1 ratio. The inclusion criteria for retrospective data were: 1) clear and complete chest X-ray images, and 2) definitive imaging diagnosis conclusions; the exclusion criteria were: 1) CXR images with severe quality issues, and 2) incomplete clinical medical records. The baseline characteristics of patients in the retrospective study are presented in Supplementary Table 1. For the prospective study, data were sourced from three hospitals in China, with a total of 296 patients enrolled in the AI-radiologist collaboration. The baseline characteristics for the prospective study are provided in Supplementary Table 2. For prospective clinical validation, the inclusion criteria included: 1) clinically suspected thoracic diseases (such as pneumonia, tuberculosis, or lung cancer) requiring CXR-assisted diagnosis, 2) patients providing written informed consent for research data use, 3) complete clinical records (including chief complaints, medical history, and laboratory test results), 4) patients with no historical chest X-ray images and no need for comparison with previous chest X-ray images, and 5) patients who underwent only posteroanterior chest X-rays without lateral chest X-rays; the exclusion criteria were: 1) substandard CXR image quality (including severe motion artifacts, over-/underexposure, or missing anatomical structures), and 2) pregnant or lactating women. In the prospective study, all included CXR images of cases were checked by radiologic technologists according to preset standards (such as no obvious motion artifacts, correct body position, and normal exposure) to ensure that the quality of the enrolled images meets the diagnostic requirements.

**Automated report generation metrics evaluation of report quality**

The fine-tuned model, Janus-Pro-CXR, was evaluated against other state-of-the-art chest radiograph interpretation models[12,24,25,27,34-36,46-50] using metrics for automated report generation. The assessment employed the RadGraph F1 score to evaluate the performance of the automated report generation system. Due to the performance variations of the CheXbert annotation tool[51] on the CXR-27 dataset, we have developed an annotation tool based on DeepSeek (Supplementary Appendix 5). Additionally, the CheXbert and DeepSeek annotation tools were employed to calculate the micro-average and macro-average F1 scores for 14 types of findings (F1-14) and the top 5 most common findings (F1-5), assessing the model's ability to identify key imaging findings. Detailed methods for these evaluations are outlined in Supplementary Appendix 6.

**Subjective Evaluation of Report Quality**

Subjective evaluation was performed by a team of five radiologists, each with 8 to 15 years of experience. All evaluators received the necessary training before participating in the evaluation (Supplementary Appendix 7). For each case, the evaluators were blinded to which report was generated by the multimodal large language model. A total of 300 cases were randomly selected from the retrospective CXR-27 test set for subjective assessment. The following evaluation methods were used: (1) Confusion Matrix: Without considering content accuracy, the order of reports generated by the large model and the published reports was randomized. The confusion matrix was then used to assess the professionalism of medical terms, logical coherence, diagnostic suggestions, and other report aspects3. (2) Report Quality Score: Evaluators rated the multimodal large language model's report generation capability against the published imaging reports, using a five-point Likert scale (5 representing the highest quality, 1 the lowest). The specific criteria for scoring are provided in Supplementary Appendix 8. (3) Pairwise Preference Test: Each evaluator was required to select the superior report from the reports generated by the multimodal large language model and the published reports, with no ambiguous responses allowed. (4) Agreement Evaluation: The RADPEER scoring system[52] was employed to assess the agreement between the original and generated reports, including the clinical significance of any differences. Detailed RADPEER scoring criteria are found in Supplementary Appendix 9. Reports used as references for subjective evaluation must undergo review, revision, and signing by senior physicians before being officially released.

**Evaluation of the collaboration between AI and radiologists**

The generation and release of radiology diagnostic reports followed a stringent standardized workflow. Junior physicians conducted a thorough review and analysis of the imaging data to draft initial diagnostic reports. These reports were then reviewed by senior physicians for accuracy and compliance with diagnostic standards. Upon verification, the reports were electronically signed by senior physicians and officially released, with real-time synchronization to the hospital information system for clinical use. Patients could access their diagnostic results through various channels, including hospital apps, self-service terminals, or printed reports. Currently, the Janus-Pro-CXR system operates independently from the radiologist review system and is deployed on a separate computer. The two systems are connected via a local area network, which enables the transmission of chest X-ray images and basic patient history to the Janus-Pro-CXR system. Upon receiving the data, the Janus-Pro-CXR system generates a radiology report. This report is then sent back to the workstation running the radiologist review system, where radiologists copy and paste it into the radiologist review system. The entire process, from data transmission to the final copy-and-paste operation, takes approximately three seconds. Radiologists can then perform any necessary edits.

In the prospective study, 20 radiologists were randomly assigned to either the AI-assisted group or the standard care group in a 1:1 ratio (Supplementary Appendix 10). Randomization was conducted using a computer-generated sequence with sealed envelopes. In the AI-assisted group, junior radiologists (residents with 1–3 years of seniority) used the reports generated by the multimodal large language model as a reference and had the option to modify them as needed. In contrast, radiologists in the standard care group completed the reports independently. For each patient, two junior radiologists from different groups independently reviewed the images and produced two separate reports. These reports were then reviewed and revised by senior radiologists before being finalized for patient delivery. The quality of reports generated by the two groups of junior radiologists was assessed using report quality scores, pairwise preference tests, and agreement evaluations. Additionally, the time taken to write reports (reading time) was compared to evaluate the impact of the multimodal large language model on work efficiency. Reading time was defined as the period from when the junior radiologists began examining the CXR until the final report was completed. This time was automatically recorded by the system. During the image interpretation process, junior radiologists had access to the full clinical medical history and prior imaging results of the patients in real time. To simulate real clinical conditions, key patient medical history information was incorporated into the Prompts for the multimodal large language model.

**Outcomes**

The primary outcomes of this study included report quality scores, pairwise preference tests, agreement evaluations, and reading time for the radiology reports produced by the two groups of junior radiologists in the prospective study. Secondary outcomes comprised the report quality scores, pairwise preference tests, and agreement evaluations comparing the published radiological reports with those generated by the model in the retrospective study.

**Validation of Multi-Image Input Capability of the Model**

To verify the model's multi-image input capability, an additional 50 patients with both posteroanterior and lateral chest radiographs, as well as 50 patients with historical chest radiographs, were included in the retrospective study for evaluation. In the prospective study, another 50 patients with posteroanterior and lateral chest radiographs were enrolled for validation. The evaluation methods included report quality scores, agreement scores, and reading time.

**Statistical analysis**

Sample size determination for the prospective study was based on the pre-experiment results regarding report quality scores (power=0.90, Supplementary Appendix 10). Paired t-tests were conducted to compare differences between the two groups, while repeated measures analysis of variance (ANOVA) was used to assess differences across multiple groups. The 95% confidence intervals were constructed using a parametric method based on the t-distribution, with the calculation formula: $\bar{x} \pm t_{\alpha/2} \times SE$, where the standard error (SE) was calculated from the sample standard deviation and sample size. The model's classification performance for various chest diseases was evaluated using the ROC curve and the AUC, with probability thresholds ranging from 0 to 1. Cohen's kappa coefficient was employed to assess the agreement between the model's results and the reference standard. Kappa values range from -1 to 1, with a value of 1 indicating perfect agreement, 0 representing agreement due to chance, and -1 signifying perfect disagreement. The Kendall's W concordance coefficient was calculated to evaluate the consistency of results across multiple raters. All statistical analyses were performed using two-sided tests, with a significance level set at $P < 0.05$. Statistical analysis was performed using R software (version 4.2.1).

**Acknowledgements:** We extend our sincere gratitude to Dr. David Ouyang, MD (Assistant Professor, Department of Cardiology, Division of AI in Medicine, Cedars-Sinai Medical Center, Los Angeles, California, USA) for his invaluable feedback and insightful suggestions during the revision of this manuscript.

**Data availability statement:** All the data supporting the findings of this study are available within this article and its Supplementary Information. Any additional information can be obtained from corresponding authors upon request. Source data are provided with this paper.

**Code availability statement:** The code for the model and the DeepSeek labeling tool can be available from the URL (https://github.com/ZrH42/Janus-Pro-CXR).

**Contributions**

Y.B., R.Z., Y.L., X.D., D.T., Y.X. C.Z., H.Z., and B.D. had major involvement in study conception or design. J.Y., S.J., C.W., W.Y., Y.G., G.Z., C.W. Q.Y., L.C., W.T., B.Z., X.W., T.S., and W.Z. had substantial involvement in data acquisition. Y.B. and Y.L. had major involvement in data analysis. Y.B., W.Y., Y.G. and G.Z. had major involvement in data interpretation. All authors had access to the data and participated in writing, reviewing, and revising the manuscript. All authors approved the final manuscript for publication. All authors accept responsibility for the accuracy and integrity of all aspects of the research.

**Conflict of Interest Disclosures:** The authors declared no conflicts of interest.

**Funding/Support:** This work was funded by the National Natural Science Foundation of China (Grant No. 62225113; 82472070), the National Key Research and Development Program of China (Grant No. 2023YFC2705700; 2023YFC2413500), the National Natural Science Foundation of China (Grant No. U25A20443; 82203144), and the Natural Science Foundation of Hubei Province of China (Grant No. 2023AFB1083 and 2025DJA055).

**Table 1. Differences in primary outcome measures of the prospective study.**

|  | Standard care | AI-assisted | P value |
| --- | --- | --- | --- |
| Report Quality Score | 4.12±0.80 | 4.36±0.50 | <0.001 |
| Agreement Score | 4.14±0.84 | 4.30±0.57 | <0.001 |
| Reading time | 147.6±51.1 | 120.6±45.6 | <0.001 |

Data were presented as mean ± SD where appropriate. Paired t-test was used to analyze the differences between the two groups (n=296).

**Figure legends**

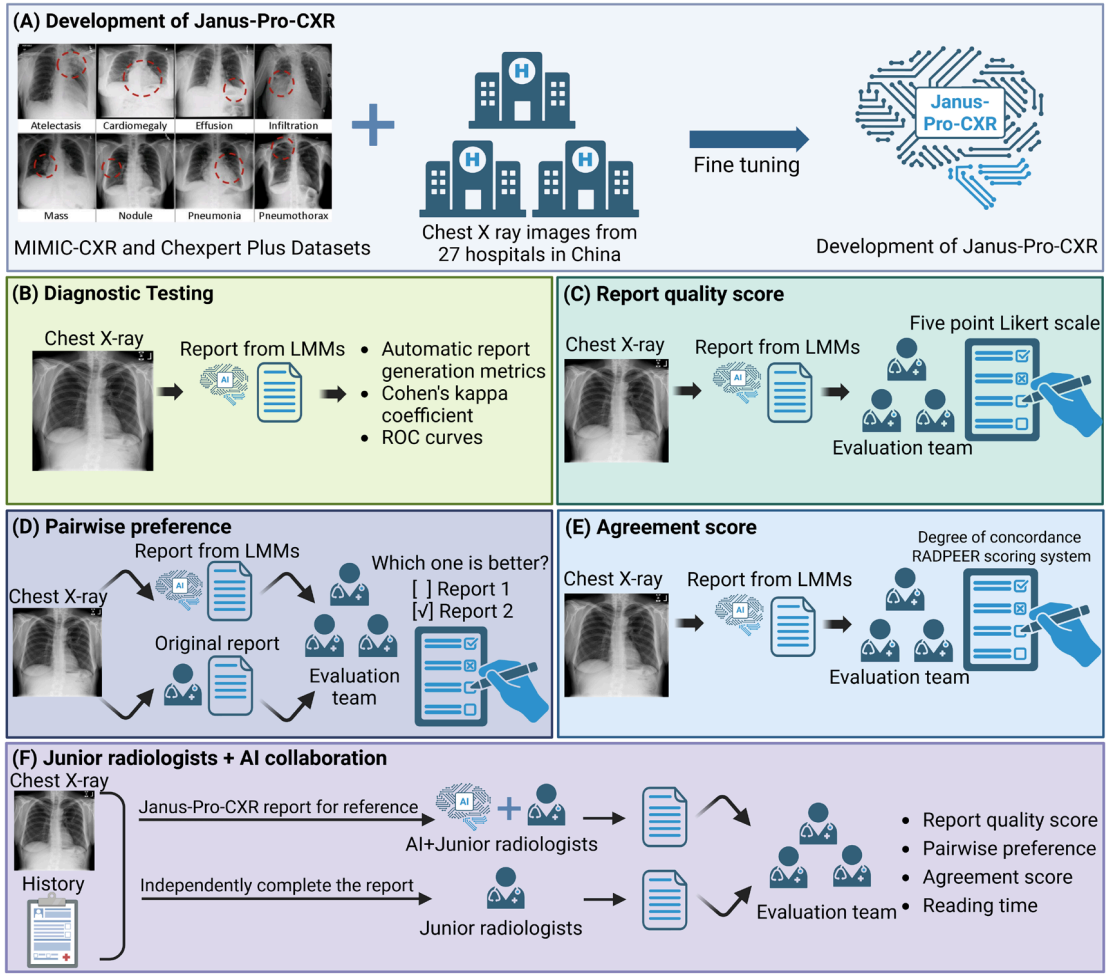

**Figure 1.** Flow chart of the systematic evaluation in this study. (A) Janus-Pro-CXR was constructed through supervised fine-tuning using the MIMIC-CXR, CheXpert Plus, and CXR-27 datasets. The retrospective data were sourced from 27 medical centers in China, collectively referred to as CXR-27. (B) The model's performance was assessed using automated report generation metrics. The quality of the generated reports was evaluated through report quality scores (C), preference tests (D), and agreement evaluations (E). (F) The clinical utility of the model was evaluated through clinical collaboration with radiologists. The figure was created using BioRender (https://biorender.com/).

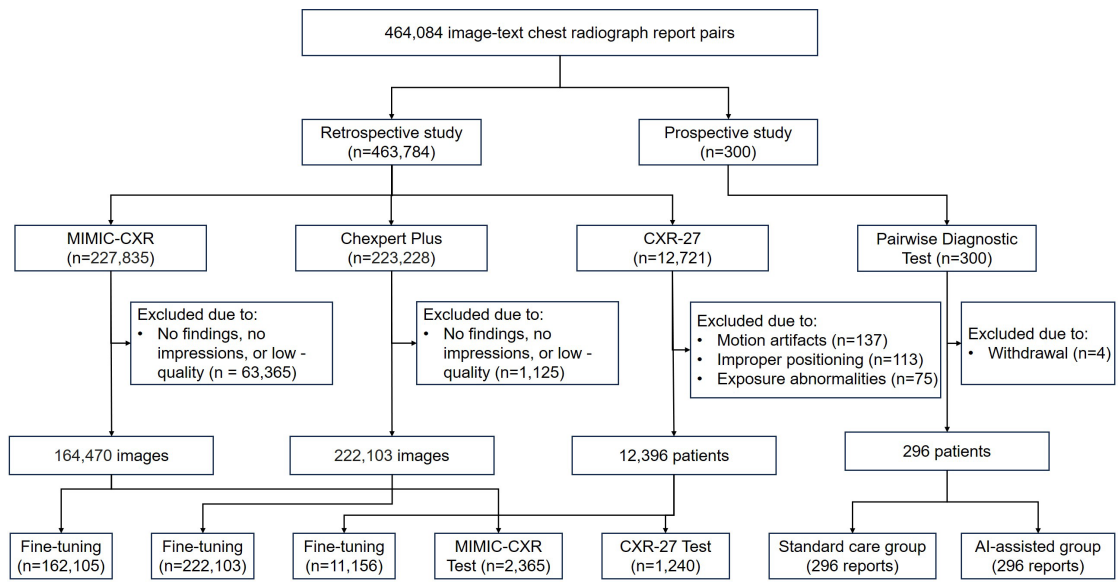

**Figure 2.** Flow chart of patient inclusion and exclusion. Janus-Pro-CXR was constructed through supervised fine-tuning using the MIMIC-CXR, CheXpert Plus, and CXR-27 datasets. The retrospective data were sourced from 27 medical centers in China, collectively referred to as CXR-27.

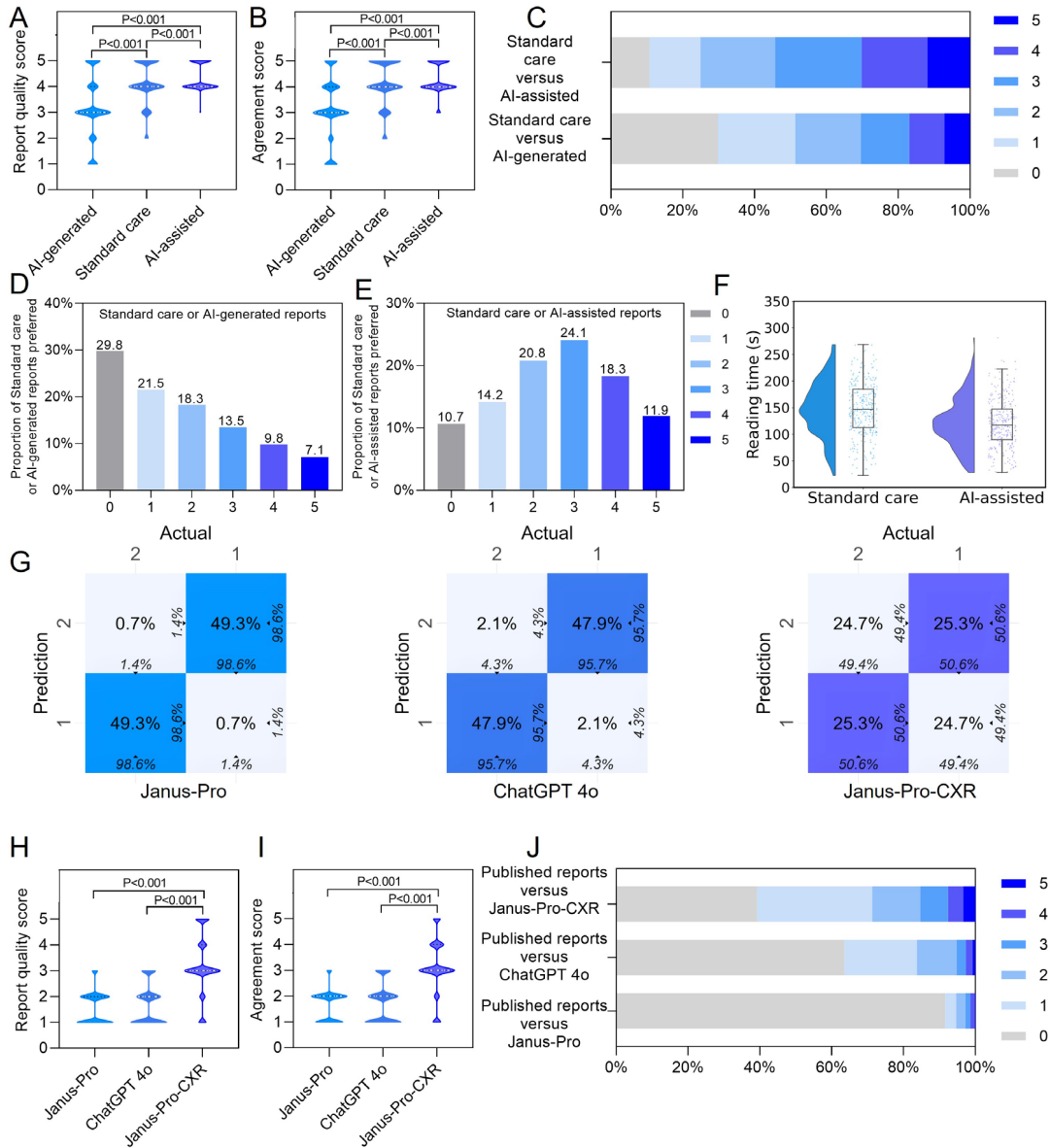

**Figure 3.** Subjective evaluations of the prospective and retrospective studies. (A) Report quality scores for the AI-generated group, the standard care group, and the AI-assisted group in the prospective study (n=296). (B) Report agreement scores for the AI-generated group, the standard care group, and the AI assisted group in the prospective study (n=296). Repeated measures ANOVA determined statistical significance for (A) and (B). Each violin's width reflects data point density at respective values; white dashed lines show the median, black dashed lines the interquartile range (25th–75th percentiles). (C) Preference tests comparing reports generated by junior radiologists and those generated by AI, as well as those generated by junior radiologists with AI collaboration, in the prospective study. (D) Preference test comparing reports generated by junior radiologists and AI.

(E) Preference test comparing reports generated by junior radiologists with AI collaboration. (F) Reading time for the standard care group and the AI assisted group in the prospective study (n=296). The paired t-test was used and the center line corresponds to the median, and the box is delineated by the first and third quartiles. (G) Confusion matrix for evaluators to identify reports generated by large models in the retrospective study. (H) Report quality scores of the three multimodal large language models in the retrospective study (n=300). (I) Report agreement scores of the three multimodal large language models in the retrospective study (n=300). Repeated measures ANOVA determined statistical significance for (H) and (I). Each violin's width reflects data point density at respective values; white dashed lines show the median, black dashed lines the interquartile range (25th–75th percentiles). (J) Preference tests comparing published reports with those generated by Janus-Pro-CXR, ChatGPT 4o, and Janus-Pro in the retrospective study.

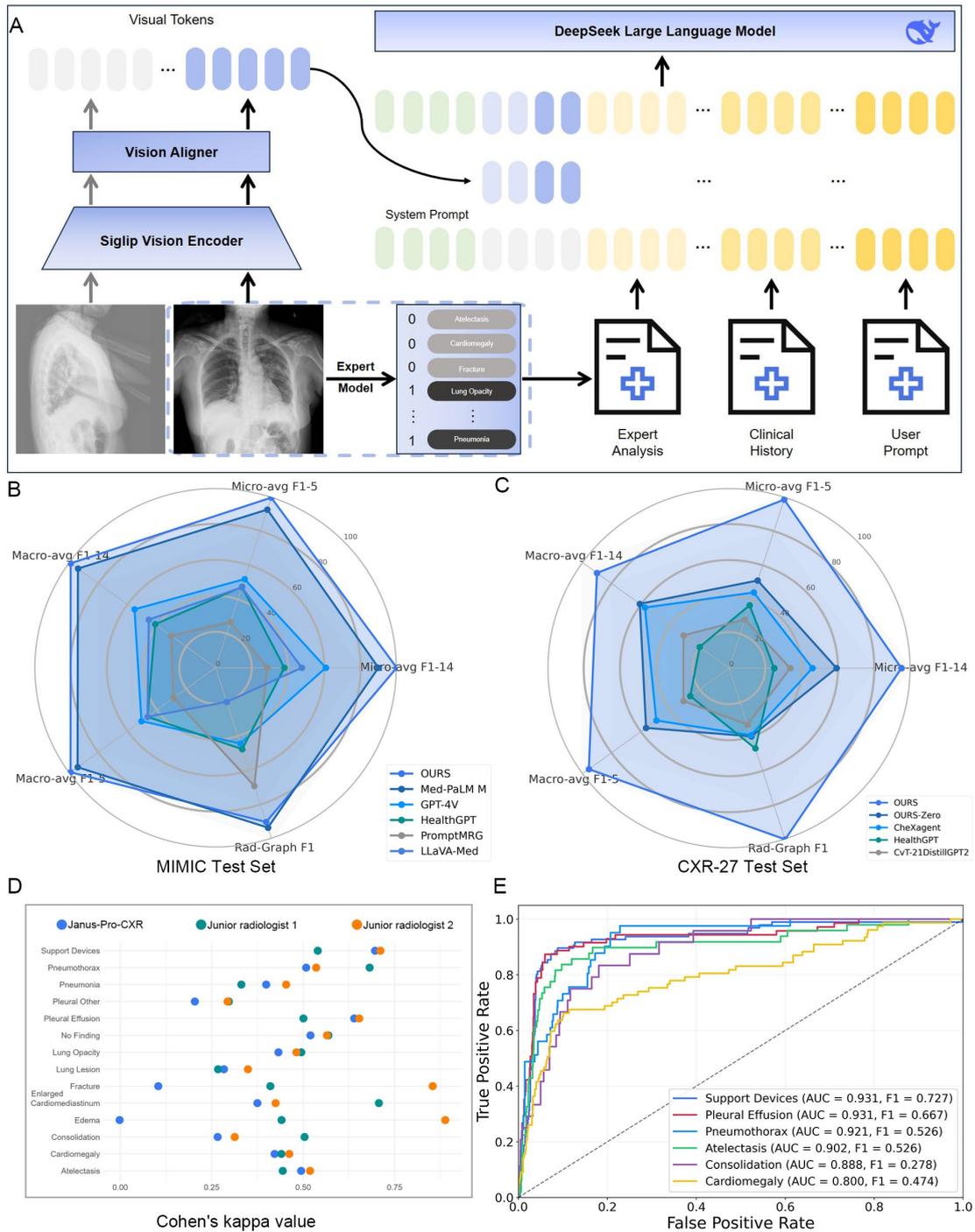

**Figure 4.** Performance of Janus-Pro-CXR on automated report generation metrics. (A) The technical architecture of the Janus-Pro-CXR system. (B) Automated report generation metrics for the MIMIC-CXR test set. Annotation was performed using the CheXbert labelling tool (uncertain labels treated as positive). The same open-source multimodal large language model was tested with parameter configurations consistent with the published research. The top 5 findings are: Atelectasis, Cardiomegaly, Edema, Consolidation, and Pleural Effusion. (C) Automated report generation

metrics for the CXR-27 test set. Annotation was performed using the DeepSeek labelling tool, and the open-source multimodal large language model was tested with consistent parameter configuration. The top 5 diseases include: Support Devices, Pleural Effusion, Lung Opacity, Pneumonia, and Lung Lesion. (D) Cohen's kappa coefficient for evaluating consistency between the model and the reference standard (n=1026). Janus-Pro-CXR-Zero refers to the model fine-tuned with the MIMIC-CXR and CheXbert Plus datasets, while Janus-Pro-CXR-Final indicates the model fine-tuned with the CXR-27 dataset, building on Janus-Pro-CXR-Zero. (E) Classification performance of the model for various chest diseases (n=1026). Evaluated through the receiver operating characteristic (ROC) curve and the area under the curve (AUC).